\documentclass[conference]{IEEEtran}
\IEEEoverridecommandlockouts

\usepackage{cite}
\usepackage{amsmath,amssymb,amsfonts}
\usepackage{algorithmic}
\usepackage{graphicx}
\usepackage{textcomp}
\usepackage{xcolor}
\def\BibTeX{{\rm B\kern-.05em{\sc i\kern-.025em b}\kern-.08em
    T\kern-.1667em\lower.7ex\hbox{E}\kern-.125emX}}
\usepackage{enumitem}
\usepackage{caption}
\usepackage{subcaption}
\usepackage{multirow}
\usepackage{makecell}
\usepackage{todonotes}
\usepackage{xcolor,colortbl}

\usepackage[dvipsnames]{xcolor}

\begin{document}


\title{Fast and Compact Tsetlin Machine Inference on CPUs Using Instruction-Level Optimization\\

\thanks{}

}

\author{
  \IEEEauthorblockN{Yefan Zeng, Shengyu Duan, Rishad Shafik, Alex Yakovlev}
  \IEEEauthorblockA{
    \textit{Microsystems Research Group, Newcastle University} \\
    Newcastle upon Tyne, United Kingdom \\
    \{y.zeng15, shengyu.duan, rishad.shafik, alex.yakovlev\}@newcastle.ac.uk
  }
}

\maketitle

\begin{abstract}
The Tsetlin Machine (TM) offers high-speed inference on resource-constrained devices such as CPUs. Its logic-driven operations naturally lend themselves to parallel execution on modern CPU architectures. Motivated by this, we propose an efficient software implementation of the TM by leveraging instruction-level bitwise operations for compact model representation and accelerated processing.
To further improve inference speed, we introduce an early exit mechanism, which exploits the TM's AND-based clause evaluation to avoid unnecessary computations. Building upon this, we propose a literal Reorder strategy designed to maximize the likelihood of early exits. This strategy is applied during a post-training, pre-inference stage through statistical analysis of all literals and the corresponding actions of their associated Tsetlin Automata (TA), introducing negligible runtime overhead.
Experimental results using the gem5 simulator with an ARM processor show that our optimized implementation reduces inference time by up to 96.71\% compared to the conventional integer-based TM implementations while maintaining comparable code density.


\end{abstract}


\section{Introduction}
Edge devices are constrained by power, memory, and latency, which challenges conventional deep neural networks (DNNs) reliant on costly floating-point operations \cite{wang2020convergence}. This has led to the exploration of alternative models that combine low complexity with logical transparency.

The TM is a rule-based machine learning model whose core feature is its complete reliance on logic-based operations rather than traditional arithmetic-intensive computations \cite{granmo2018tsetlin}. Compared to multi-layer neural networks (NNs), the TM can achieve higher classification accuracy in certain tasks, while its reliance solely on logical operations, as opposed to thousands of multiply-accumulate operations in NNs, results in significantly lower computational complexity \cite{duan2025ethereal}. It constructs propositional logic clauses from Boolean literals using many simple learning agents called Tsetlin Automata (TA). This structure has extremely high interpretability, as each clause can be directly read as a human-understandable logical rule.

\begin{figure}[t]
  \centering
  \includegraphics[width=0.66\linewidth]{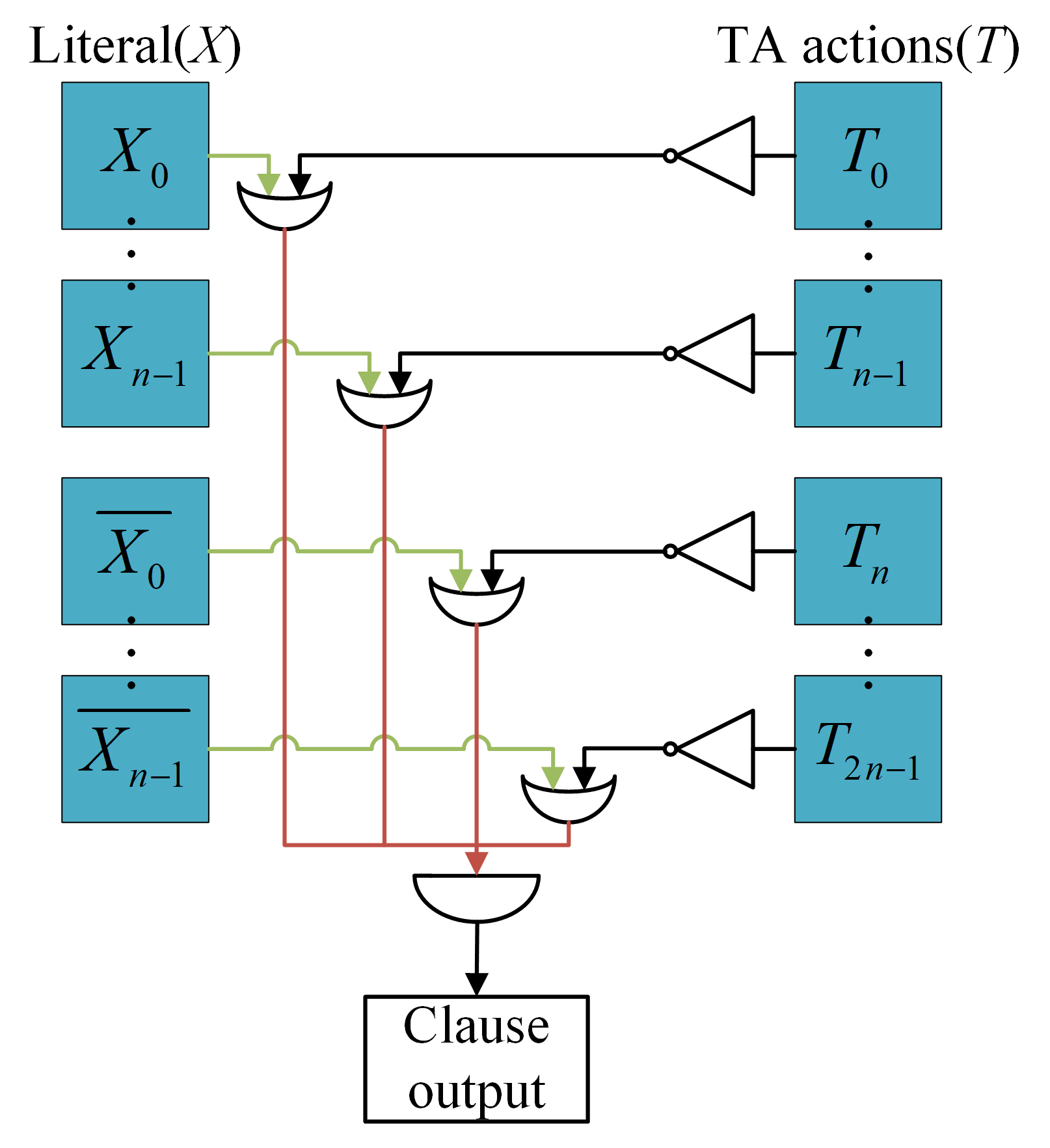}
   \caption{TM inference clause output}
    \label{fig:TM inference clause output}
\end{figure}

\begin{figure}[t]
    \centering 
    \includegraphics[width=1\columnwidth]{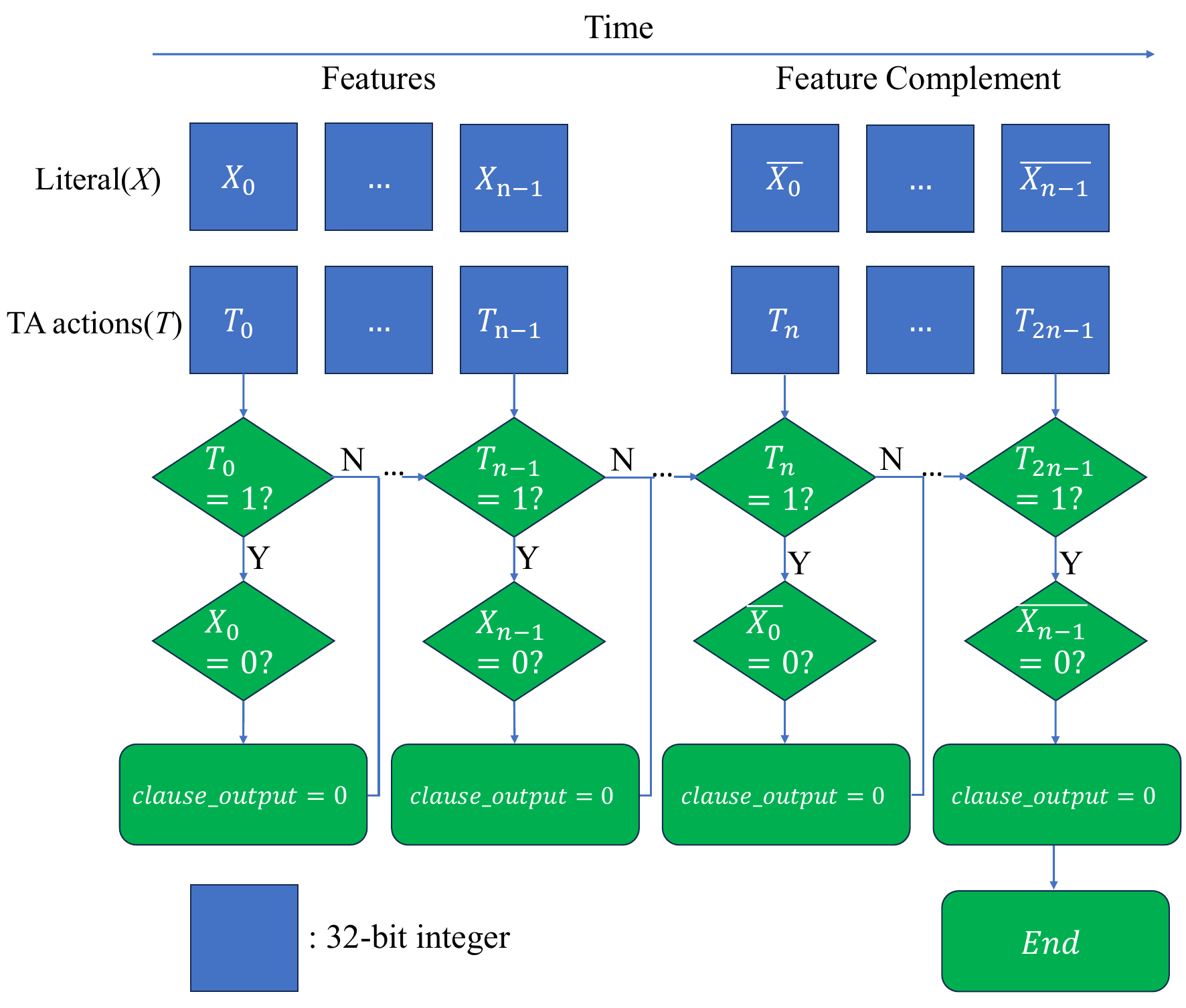} 
    \caption{Integer operation inference}
    \label{fig:Integer operation inference}
\end{figure}

Given that the TM is fundamentally a logic-based machine learning model with low computational complexity, it is inherently well-suited for execution on less powerful general-purpose processors like CPUs. This motivates the exploration of lightweight software implementations. However, existing TM inference designs are often realized using integer-based operations and conditional branching \cite{granmo2018tsetlin}. While functional, such implementations do not align with the logical nature of TM, resulting in unnecessary overhead and inflated model representation, as shown in Fig.~\ref{fig:TM inference clause output} and Fig.~\ref{fig:Integer operation inference}, where n represents the feature index.

To address this gap, REDRESS \cite{maheshwari2023redress} proposed a compressed, bit-level TM inference method that significantly reduces runtime and memory usage and better matches the logic-gate interpretation of TM clause evaluations. While REDRESS represents an important step forward, it is constrained to post-training inference, meaning inference can only begin once training is complete and compressed models are finalized. This hinders its applicability to real-time computing scenarios, which are especially critical in edge applications—where TM is commonly deployed for classification tasks under strict latency constraints \cite{11105163}.

We explored the fact that TM clause computation is inherently based on the AND operation among a large number of Boolean literals. This logical structure naturally enables the implementation of early exit, as a single literal with value 0 will immediately determine the clause output. Moreover, since both TA actions and literals are represented in Boolean form, the inference process is well-suited for bitwise operations, offering significant computational advantages.
To further enhance the efficiency of early exit, we introduce a Reorder strategy performed at post-training, pre-inference stage, aiming to position high-impact literals earlier in the evaluation sequence and thus increase the likelihood of early exit.

\section{TM Instruction-level Optimization}

\subsection{Bitwise operation and early exit}
\begin{figure}[htbp]
    \centering
    \includegraphics[width=1\linewidth]{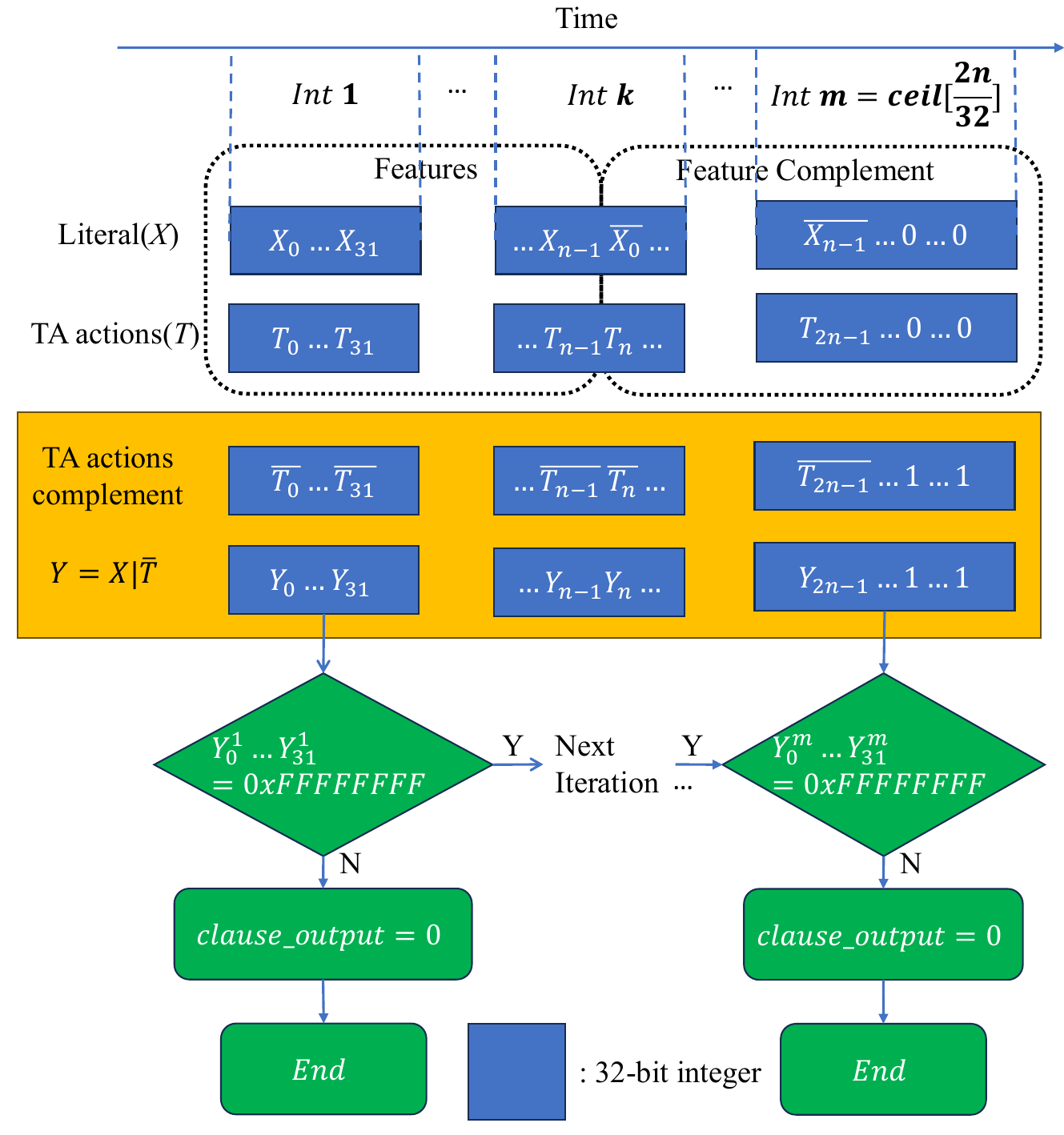}
    \caption{Bitwise operation with early exit}
    \label{fig:Bitwise operation with early exit}
\end{figure}

After analyzing and observing the performance bottlenecks in the integer-based TM inference process, We propose a more efficient inference strategy that utilizes pure integer-based logic operations to optimize clause output computation, building upon the open-source Vanilla TM implementation \cite{yordzhev2013bitwise, tsetlinmachinec}. We precisely replicate this logic behavior using bitwise operations. This approach not only aligns more closely with the theoretical essence of TM but also ensures logical consistency for subsequent hardware implementations (e.g., FPGA or ASIC), as shown in Fig.~\ref{fig:Bitwise operation with early exit}.

In terms of bitwise operation design, we uniformly adopt a 32-bit integer-based loading method, dividing the original input into \( m = \left\lceil \frac{2n}{32} \right\rceil \) integer units. If the last group of literals is less than 32 bits, it is padded with zeros. Similarly, the TA actions file is also padded in this manner when loaded. It should be noted that such padding does not affect the correctness of the inference: in the TA actions file, the padded part is defaulted to the exclude state; according to the TM inference logic, if a literal is marked as exclude, even if its value is arbitrary, it has no effect on the clause output (it is always logical 1 in an OR gate).

Subsequently, we perform a complement operation on the TA actions and perform a bitwise AND operation with the corresponding literal. we adopt a more streamlined bit-level processing approach: on the timeline, we iterate from left to right in integer order (\(\mathit{int}\ 1\)
 to \( \mathit{int}\ m = \left\lceil \frac{2n}{32} \right\rceil \)
), and in each round, map 32 literals and 32 TA actions to the 32 bits of an integer, performing the AND operation between the corresponding literal and complement (TA actions).

At the end of each iteration, we check equal to \(
0\text{x}FFFFFFFF
\)
 to determine if there are conditions that make the clause output 0. Once a non-all-1 result is detected, the early exit logic is triggered, immediately setting the clause output to 0; otherwise, the process proceeds to the next iteration until all integers are processed. After completing the clause output calculation, the process enters the class sum voting phase, consistent with the integer-based TM inference process.

We observed that in the integer-based TM implementation, as shown in Fig.~\ref{fig:Integer operation inference}, the early exit mechanism is not used. Even when \(\mathit{clause\_output} = 0\) occurs, the program continues to execute until the last literal is reached. To address this efficiency issue, we designed an early exit strategy: once \(\mathit{clause\_output} = 0\) is detected, the loop is immediately exited. In our design, this mechanism is implemented in each iteration by checking whether any bit of the current 32-bit integer is 0. If a 0 is detected, the loop is immediately exited, thereby achieving early exit and avoiding unnecessary computational overhead.

\subsection{Reorder and Booleanization}
During training, we observed that the number of states in the TA with a value of 1 (i.e., include) was far less than the number with a value of 0 (i.e., exclude). However, early exit can only be triggered when a literal is marked as include and the input does not satisfy the condition. Therefore, we aim to explore whether the newly designed inference process can achieve a break within the first few iterations by reordering literals in datapoints and adjusting the order of TA actions, accordingly, thereby improving execution efficiency.

\begin{figure}
    \centering
    \includegraphics[width=1\linewidth]{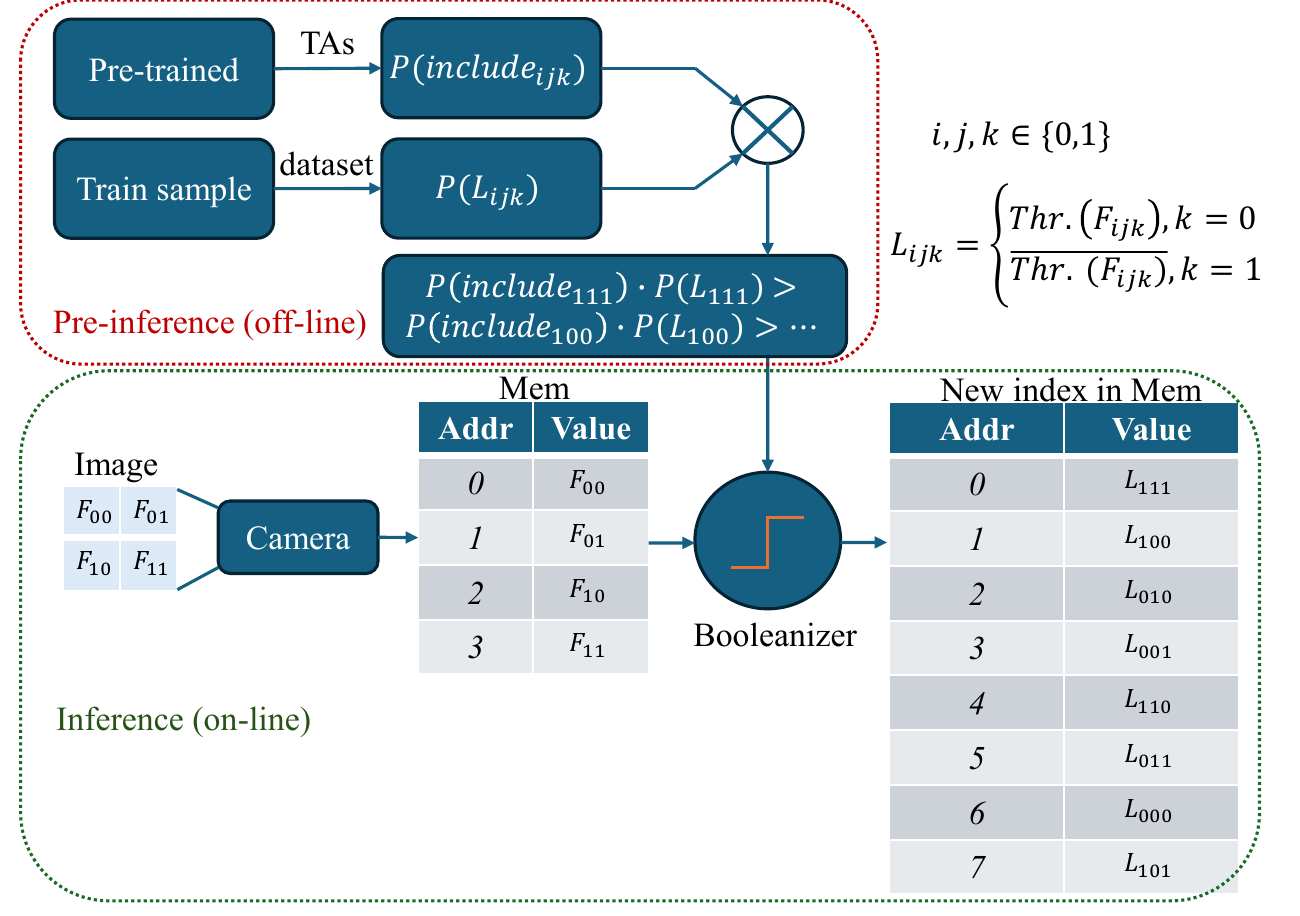}
    \caption{Booleanization with Reorder index}
    \label{fig:Booleanization with Reorder index}
\end{figure}

To begin with, the raw features extracted from an image are denoted as \( F_{ij} \). Due to the involvement of complemented literals in the TM, a parameter \( k \) is introduced to distinguish between original and negated forms. After the Booleanization process, each feature \( F_{ij} \) is converted into a literal \( L_{ijk} \), which may represent either the original value or its complement depending on \( k \), as shown in Fig.~\ref{fig:Booleanization with Reorder index}.

To determine the order of literals when reordering, we introduce the following metrics as Reorder criteria: First, we calculate the probability that each literal is 0 in all datapoints, denoted as \( P(L_{ijk}) \); simultaneously, we compute the probability that the literal corresponds to TA action = 1 (i.e., included) in all clauses, denoted as \( P(include_{ijk}) \). Next, we multiply \( P(L_{ijk}) \) by \( P(include_{ijk}) \) for each literal, sort the results in descending order based on the product, and finally obtain the new index order to adjust the arrangement of input literals and TA actions, as shown in \eqref{eq:addr_order}. 

\begin{equation}
\begin{split}
P(include_a)\cdot P(L_a) &> P(include_b)\cdot P(L_b) \\
&\implies \text{Addr}(L_a) < \text{Addr}(L_b)
\end{split}
\label{eq:addr_order}
\end{equation}

where $\forall L_a, L_b \in \{L_{ijk}\}$.

In design concept, Reorder and Booleanization are two highly coupled and synergistic processes. Reorder strategy can be regarded as an intermediate stage between training and inference. In the real-time computing workflow, the TA actions data obtained during the training phase is stored in virtual memory, so there is no need to localise it for conventional post-training inference. When new raw features from sensors like cameras are collected, they can be directly initialised and written to memory. Subsequently, the Reorder strategy can be performed in this phase and reschedule the Booleanization process based on the order of the obtained Reorder index.

In summary, if Reorder strategy is not performed, the system still must undergo a process of reading raw features from memory, performing Booleanization processing, and then writing them back to memory. This opens a key direction for future work—examining whether the Reorder strategy introduces overhead and if its efficiency gains during inference can outweigh this cost, which is vital for real-time edge deployment.

\section{Experiment and results}
\subsection{Experimental setup}
In this study, we adopted the gem5 architecture-level simulator to significantly enhance experimental efficiency. Compared to the time-consuming and inflexible preparation required for deployment on actual hardware or microcontroller units (MCUs), gem5 offers a flexible and highly controllable simulation environment, allowing for rapid design iteration and configuration testing under varying scenarios \cite{binkert2011gem5}.

After successfully setting up the simulator and its dependencies, we selected a widely used commercial ARM processor architecture as the baseline instruction set. Although gem5 is a simulator and does not represent real hardware directly, in subsequent experiments, we use technical specifications of representative STM32F746G-DISCO platform (e.g., 216 MHz clock frequency) as reference benchmarks to better approximate practical deployment conditions.

Since the current work focuses solely on the inference stage, we employ post-training datasets and corresponding TA actions to conduct performance evaluation. We used Iris \cite{iris_53} and MNIST \cite{deng2012mnist} dataset configurations, a Booleanization step is initially performed for each dataset to transform raw features into corresponding Boolean representations \cite{rahman2022data}, as shown in TABLE~\ref{Dataset and TM model details}.

\begin{table}[htbp]
\centering
\caption{Dataset and TM model details}
\label{Dataset and TM model details}
\resizebox{\linewidth}{!}{%
\begin{tabular}{|c|c|c|c|c|c|c|}
\hline
Dataset & Classes & Features & Literals & Clauses\textsuperscript{a} & \((T, s)\) & Test Acc. (\%) \\ \hline
Iris & 3 & 48 & 96 & 16 & (8,4) & 94.67 \\ \hline
\multirow{2}{*}{MNIST} & \multirow{2}{*}{10} & \multirow{2}{*}{784} & \multirow{2}{*}{1568} & 20 & (3,10) & 90 \\ \cline{5-7}
 & & & & 100 & (10,8) & 95.70 \\ \hline
\end{tabular}%
}
\vspace{1mm}
\begin{flushleft}
\textsuperscript{a} Number of clauses per class
\end{flushleft}
\end{table}

\subsection{Bitwise operation and early exit}

Under these configurations, we introduced our optimized method and compared it with the integer-based TM inference implementation (based on integer judgment logic). The design method employs bitwise operations to generate clause outputs and incorporates an early exit mechanism to optimize runtime efficiency.

The final experimental results are shown in TABLE~\ref{Results of bitwise operation, early exit and code density}, illustrating the performance differences between different implementation methods across multiple datasets and configurations.

\begin{table*}[]
\caption{Results of bitwise operation, early exit and code density}
\label{Results of bitwise operation, early exit and code density}
\resizebox{\linewidth}{!}{%
\begin{tabular}{|c||c|c||c|c|c||c|c|c||c|c|c|}
\hline

\multirow{2}{*}{} & \multicolumn{2}{c||}{\textbf{Baseline}} & \multicolumn{3}{c||}{\textbf{Early exit only}} & \multicolumn{3}{c||}{\textbf{Bitwise only}} & \multicolumn{3}{c|}{\textbf{Early exit + Bitwise}} \\  \cline{2-12}
& \begin{tabular}[c]{@{}c@{}} Inference\\ time\end{tabular} & \begin{tabular}[c]{@{}c@{}} Code\\ density (kB)\end{tabular} & \begin{tabular}[c]{@{}c@{}} Inference\\ time\end{tabular} & \cellcolor{Yellow!50!white} \begin{tabular}[c]{@{}c@{}}Time\\ reduct. (\%)\end{tabular} & \begin{tabular}[c]{@{}c@{}} Code\\ density (kB)\end{tabular} & \begin{tabular}[c]{@{}c@{}} Inference\\ time\end{tabular} & \cellcolor{LimeGreen!50!white} \begin{tabular}[c]{@{}c@{}}Time\\ reduct. (\%)\end{tabular} & \begin{tabular}[c]{@{}c@{}} Code\\ density (kB)\end{tabular} & \begin{tabular}[c]{@{}c@{}} Inference\\ time\end{tabular} & \cellcolor{GreenYellow!50!white} \begin{tabular}[c]{@{}c@{}}Time\\ reduct. (\%)\end{tabular} & \begin{tabular}[c]{@{}c@{}} Code\\ density (kB)\end{tabular} \\ \hline
\begin{tabular}[c]{@{}c@{}}Iris\\ (16 Clause)\end{tabular} & 68 ms & 354.23 & 26.39 ms & \cellcolor{Yellow!50!white} 61.19 & 354.23 & 18.75 ms & \cellcolor{LimeGreen!50!white} 72.43 & 354.12 & 9.35 ms & \cellcolor{GreenYellow!50!white} 86.25 & 354.06  \\ \hline
\begin{tabular}[c]{@{}c@{}}MNIST\\ (20 Clause)\end{tabular} & 3.09 s & 354.23 & 1.43 s & \cellcolor{Yellow!50!white} 53.72 & 354.23 & 0.28 s & \cellcolor{LimeGreen!50!white} 90.94 & 354.36 & 0.19 s & \cellcolor{GreenYellow!50!white} 93.85 & 354.36  \\ \hline
\begin{tabular}[c]{@{}c@{}}MNIST\\ (100 Clause)\end{tabular} & 15.22 s & 354.23 & 8.22 s & \cellcolor{Yellow!50!white} 45.99 & 354.23 & 1.21 s & \cellcolor{LimeGreen!50!white} 92.05 & 354.36 & 0.79 s & \cellcolor{GreenYellow!50!white} 94.81 & 354.36  \\ \hline
\begin{tabular}[c]{@{}c@{}}Average\\ time reduct.(\%)\end{tabular} & \multicolumn{3}{c|}{\cellcolor{gray!70!white}} & \cellcolor{Yellow!50!white} \textbf{53.63} & \multicolumn{2}{c|}{\cellcolor{gray!70!white}} &  \cellcolor{LimeGreen!50!white} \textbf{85.14} & \multicolumn{2}{c|}{\cellcolor{gray!70!white}} & \cellcolor{GreenYellow!50!white} \textbf{91.64} & \cellcolor{gray!70!white} \\ \hline
\end{tabular}
}
\end{table*}

TABLE~\ref{Results of bitwise operation, early exit and code density} compares the integer-based TM inference implementation with the optimized method proposed in this study, which employs bitwise operations and an early exit mechanism. As shown in the results, our method significantly reduces the total inference latency across all dataset configurations, with the most notable improvement observed when both bitwise operations and early exit are enabled. The latency reduction reached 86.25\%, 93.85\%, and 94.81\% for the respective datasets, while maintaining comparable code density. This advantage is particularly pronounced in datasets with a large number of features (e.g., MNIST), suggesting that the proposed method offers greater potential for efficiency improvements in high-dimensional input scenarios.

\subsection{Reorder}
By comparing different datasets, we evaluated the impact of the Reorder strategy against the baseline implementation, as illustrated in the TABLE~\ref{Results of Reorder strategy}. Experimental results demonstrate that the utilization of Reorder strategy, on top of bitwise operations and early exit, leads to further performance gains—specifically, inference time was reduced by 87.68\%, 95.47\%, and 96.71\% for the respective datasets. The results also suggest that Reorder strategy yields more significant improvements on larger datasets. Further experiments will be conducted to explore whether the Reorder strategy itself can be further optimized.

\begin{table}[htbp]
\caption{Results of Reorder strategy}
\label{Results of Reorder strategy}
\resizebox{\linewidth}{!}{%
\begin{tabular}{|c||c||c|c||c|c|}
\hline
\multirow{2}{*}{} & \textbf{Baseline} & \multicolumn{2}{c||}{\textbf{w/o. Reorder}} & \multicolumn{2}{c|}{\textbf{Full optimization (w. Reorder)}} \\  \cline{2-6}
& \begin{tabular}[c]{@{}c@{}} Inference\\ time\end{tabular} & \begin{tabular}[c]{@{}c@{}} Inference\\ time\end{tabular} & \begin{tabular}[c]{@{}c@{}}Time\\ reduct. (\%)\end{tabular} & \begin{tabular}[c]{@{}c@{}} Inference\\ time\end{tabular} & \begin{tabular}[c]{@{}c@{}}Time\\ reduct. (\%)\end{tabular} \\ \hline
\begin{tabular}[c]{@{}c@{}}Iris\\ (16 Clause)\end{tabular} & 68 ms & 9.35 ms& 86.25 & 8.38 ms & 87.68 \\ \hline
\begin{tabular}[c]{@{}c@{}}MNIST\\ (20 Clause)\end{tabular} & 3.09 s & 0.19 s & 93.85 & 0.14 s & 95.47 \\ \hline
\begin{tabular}[c]{@{}c@{}}MNIST\\ (100 Clause)\end{tabular} & 15.22 s & 0.79 s & 94.81 & 0.5 s & 96.71 \\ \hline
\begin{tabular}[c]{@{}c@{}}Average\\ time reduct.(\%)\end{tabular} &  \multicolumn{2}{l|}{\cellcolor{gray!70!white}} & \textbf{91.64} & \multicolumn{1}{l|}{\cellcolor{gray!70!white}} & \textbf{93.29} \\ \hline
\end{tabular}
}
\end{table}

\section{Conclusion}
This paper proposed an optimized TM inference design for low-power, real-time edge computing. By adopting bitwise operations and an early exit mechanism, we significantly reduced inference latency. A Reorder strategy was also introduced to enhance early exit, and together with Booleanization, lays the foundation for real-time deployment. Experimental results on gem5 using ARM processors demonstrate up to 96.71\% reduction in inference time. The results demonstrate the high effectiveness of the proposed inference acceleration methods while preserving strong flexibility and adaptability through their software-based implementation.

\section*{Acknowledgment}
This work was supported by EPSRC EP/X036006/1 Scalability Oriented Novel Network of Event Triggered Systems (SONNETS) project and by EPSRC EP/X039943/1 UKRI-RCN: Exploiting the dynamics of self-timed machine learning hardware (ESTEEM) project.



\end{document}